# Efficient Structured Prediction with Latent Variables for General Graphical Models

**Alexander G. Schwing**                                                    ASCHWING@INF.ETHZ.CH
Computer Science Department, ETH Zurich, 8092 Zurich, Switzerland

**Tamir Hazan**                                                            TAMIR@TTIC.EDU
Toyota Technological Institute at Chicago, Chicago, IL 60637, USA

**Marc Pollefeys**                                                         POMARC@INF.ETHZ.CH
Computer Science Department, ETH Zurich, 8092 Zurich, Switzerland

**Raquel Urtasun**                                                         RURTASUN@TTIC.EDU
Toyota Technological Institute at Chicago, Chicago, IL 60637, USA

## Abstract

In this paper we propose a unified framework for structured prediction with latent variables which includes hidden conditional random fields and latent structured support vector machines as special cases. We describe a local entropy approximation for this general formulation using duality, and derive an efficient message passing algorithm that is guaranteed to converge. We demonstrate its effectiveness in the tasks of image segmentation as well as 3D indoor scene understanding from single images, showing that our approach is superior to latent structured support vector machines and hidden conditional random fields.

## 1. Introduction

In the past few years, structured models have become an important tool in many domains such as natural language processing, computer vision and computational biology. While these models typically assume a supervised setting (*i.e.*, one has access to fully labeled input-output pairs), existing applications can benefit largely from the use of weakly labeled data. In computer vision, for example, we might want to segment an image by classifying each pixel into a semantic category, however, gathering annotated data is a very expensive process (*i.e.*, it takes several minutes to annotate a single image). The use of weakly annotated data is even more important in domains such as medical diagnosis, as observing all labels might not be possible (*e.g.*, if a hospital does not have access to a particular test/procedure).

Several structured prediction frameworks have been developed to deal with weakly labeled information. The most notable examples are hidden conditional random fields (HCRFs) (Quattoni et al., 2007) and latent structured support vector machines (LSSVMs) (Yu & Joachims, 2009). Both approaches are framed as regularized surrogate loss minimization tasks, and treat the missing annotations as hidden variables. They achieve good performance when it is possible to compute the maximum a-posteriori (MAP) estimate or the partition function exactly. Unfortunately, this is only possible for a few special cases, e.g., models with sub-modular energies or models with low tree-width graphs, which are not very common in practice. To deal with cyclic graphs, HCRFs and LSSVMs usually resort to approximate inference algorithms and heuristics regarding the stopping criteria. This can, however, significantly harm their performance as guarantees regarding valid cutting planes and decreasing the cost function at each iteration are no longer possible.

In this paper we first show that HCRFs and LSSVMs are instances of a more general framework that we refer to as *structured loss minimization with latent variables*. We then construct an approximation for this general structured prediction formulation, using duality, based on a local entropy approximation and derive an efficient message-passing algorithm that is guaranteed to converge for any type of potential and graphical model structure. We demonstrate the effectiveness of our approach on a synthetic segmentation task, as well as in the challenging vision task of inferring the 3D scene layout from single images, and show that our approach significantly outperforms LSSVM in terms of performance and HCRF in terms of speed, being 35 times faster. Additionally, for the 3D scene understanding task we show that state-of-the-art results can be obtained while utilizing only a subset of the annotations used by existing approaches.

In the following, we first derive our unified framework which contains HCRFs and LSSVMs as special cases (Sec. 2). We then describe our approximation (Sec. 3) and message passing algorithm (Sec. 4), followed by our experimental evaluation and a discussion.

## 2. Loss minimization with latent variables

In this section we propose a general framework for loss minimization with latent variables. Consider the setting where $\mathcal{X}$ is the input space (*e.g.*, an image or a sentence) and $\mathcal{S}$ is a structured label space (*e.g.*, an image segmentation or a parse tree). Note that $\mathcal{S}$ can depend on the example $x \in \mathcal{X}$. For clarity of notation we dropped this dependency while noting that neither the derivation nor our implementation have this restriction. Let $\phi : \mathcal{X} \times \mathcal{S} \to \mathbb{R}^F$ denote a mapping from input and label space to an $F$-dimensional feature space. We are interested in finding the parameters $w \in \mathbb{R}^F$ of a log-linear model, which best describe the possible labelings $s \in \mathcal{S}$ of $x \in \mathcal{X}$, *i.e.*,

$$p_w(s|x) \propto \exp\left(w^\top \phi(x, s)\right). \qquad (1)$$

In this paper, we tackle the weakly supervised setting, where we are given a training set $\mathcal{D} = \{(x_i, y_i)_{i=1}^N\}$ containing $N$ pairs, each composed by an input $x \in \mathcal{X}$ and some partially labeled data $y \in \mathcal{Y} \subseteq \mathcal{S}$. For every training pair, we divide the label space $\mathcal{S} = \mathcal{Y} \times \mathcal{H}$ into two non-intersecting subspaces $\mathcal{Y}$ and $\mathcal{H}$ and refer to the missing information $h \in \mathcal{H}$ as hidden or latent.

For many applications, we can construct a loss function $\ell_{(x,y)}(s)$ which compares a configuration $s$ with the labeled data $(x, y) \in \mathcal{D}$, providing a measure for the fitness of the estimate. We incorporate this loss function in learning by considering the distribution

$$p_{(x,y)}(s|w) \propto \exp(w^\top \phi(x, s) + \ell_{(x,y)}(s)). \qquad (2)$$

Intuitively it places more probability mass on those parts of the space $\mathcal{S}$ that have a high loss, forcing the model to learn in a more difficult setting than the one encountered at inference, where the loss is not present.

A maximum likelihood approach aims at finding model parameters $w$ which assign highest probability to the data $\mathcal{D}$. As we have no information available for the unobserved space $\mathcal{H}$ we marginalize it out, *i.e.*, we average over all possible hidden states. Therefore, we define the loss-augmented likelihood of a prediction

$\hat{y} \in \mathcal{Y}$ when observing the pair $(x, y)$ as

$$p_{(x,y)}(\hat{y}|w) \propto \sum_{\hat{h} \in \mathcal{H}} p_{(x,y)}(\hat{y}, \hat{h}|w) = \sum_{\hat{h} \in \mathcal{H}} p_{(x,y)}(\hat{s}|w). \qquad (3)$$

Assuming the data to be independent and identically distributed (i.i.d.), our goal is to minimize the negative log-likelihood $-\ln[p(w) \prod_{(x,y) \in \mathcal{D}} p_{(x,y)}(y|w)]$ with $p(w) \propto e^{-\|w\|_p^p}$ being a prior on the model parameters. As a result, the negative log-likelihood is a difference of convex terms

$$\frac{C}{p}\|w\|_p^p + \sum_{(x,y) \in \mathcal{D}} \Bigg( \ln \sum_{\hat{s} \in \mathcal{S}} \exp\left(w^\top \phi(x, \hat{s}) + \ell_{(x,y)}(\hat{s})\right) - $$
$$- \ln \sum_{\hat{h} \in \mathcal{H}} \exp\left(w^\top \phi(x, (y, \hat{h})) + \ell_{(x,y)}^c((y, \hat{h}))\right) \Bigg), \qquad (4)$$

with the first two terms being the sum of the log-prior and the logarithm of the partition function. We take the loss of a ground truth configuration $\ell_{(x,y)}^c((y, \hat{h})) = \ell_{(x,y)}((y, \hat{h})) \equiv 0$, independent of any estimate $\hat{h}$. To control the variance of the log-linear probability model we follow (Hazan & Urtasun, 2010; Pletscher et al., 2010) and introduce a temperature parameter $\epsilon$, *i.e.*,

$$\frac{C}{p}\|w\|_p^p + \sum_{(x,y) \in \mathcal{D}} \Bigg( \epsilon \ln \sum_{\hat{s} \in \mathcal{S}} \exp\left(\frac{w^\top \phi(x, \hat{s}) + \ell_{(x,y)}(\hat{s})}{\epsilon}\right) - $$
$$- \epsilon \ln \sum_{\hat{h} \in \mathcal{H}} \exp\left(\frac{w^\top \phi(x, (y, \hat{h})) + \ell_{(x,y)}^c((y, \hat{h}))}{\epsilon}\right) \Bigg). \qquad (5)$$

Importantly, $\epsilon$ defines an entire family of structured prediction tasks with latent variables. For $\epsilon = 1$ we obtain the maximum likelihood (HCRF) framework, while $\epsilon = 0$ results in the max-margin formulation for latent variables (LSSVM) $\min_w \frac{C}{p}\|w\|_p^p + \max_{\hat{s}}(w^\top \phi(x, \hat{s}) + \ell_{(x,y)}(\hat{s})) - \max_{\hat{h}}(w^\top \phi(x, (y, \hat{h})) + \ell_{(x,y)}^c((y, \hat{h})))$. Note that $\epsilon \to 0$ smoothly approximates the max-function via the soft-max.

## 3. Approximate latent structured loss minimization

The unconstrained minimization problem in Eq. (5) w.r.t. $w$ is challenging as it involves a sum of convex and concave terms containing exponentially sized sums. To make the minimization more tractable, we follow Yuille & Rangarajan (2003) and upper-bound the concave part via a minimization over a set of dual variables subsequently referred to as $q_{(x,y)}(h)$. This results in a convex dual and a non-convex bi-linear term as described in the following claim.

---

**Program 1** *Approximated structured prediction with latent variables*

$$
\min_{d,\lambda,w} \; f_1 \left\{ \frac{C}{2}\|w\|_2^2 + \sum_{(x,y)\in\mathcal{D}} \left( \sum_{i\in\mathbb{S}} \epsilon c_i \ln\sum_{s_i}\exp\left(\frac{\phi_{(x,y),i}(s_i) - \sum_{\alpha\in N(i)}\lambda_{(x,y),i\to\alpha}(s_i)}{\epsilon c_i}\right) + \right. \right.
$$
$$
\left. \left. + \sum_{\alpha\in E}\epsilon c_\alpha \ln\sum_{s_\alpha}\exp\left(\frac{\phi_{(x,y),\alpha}(s_\alpha) + \sum_{i\in N(\alpha)}\lambda_{(x,y),i\to\alpha}(s_i)}{\epsilon c_\alpha}\right)\right)\right) -
$$

$$
f_2 \left\{ -\sum_r w_r \left( \sum_{(x,y)}\left(\sum_{i\in\mathbb{Y}}\phi_{r,i}(x,y_i) + \sum_{i\in\mathbb{H},h_i}\phi_{r,i}(x,h_i)d_{(x,y),i}(h_i) + \sum_{\alpha\in E,h_\alpha}\phi_{r,\alpha}(x,(y,h)_\alpha)d_{(x,y),\alpha}(h_\alpha)\right)\right)\right.
$$

$$
f_3 \left\{ \begin{aligned} &-\sum_{(x,y)}\left(\sum_{i\in\mathbb{H},h_i}\ell^c_{(x,y),i}(x,h_i)d_{(x,y),i}(h_i) + \sum_{\alpha\in E_\mathbb{H},h_\alpha}\ell^c_{(x,y),\alpha}(x,(y,h)_\alpha)d_{(x,y),\alpha}(h_\alpha)\right)\\ &-\sum_{(x,y)}\left(\sum_{i\in\mathbb{H}}\epsilon\hat{c}_i H(d_{(x,y),i}) + \sum_{\alpha\in E_\mathbb{H}}\epsilon\hat{c}_\alpha H(d_{(x,y),\alpha})\right)\end{aligned}\right.
$$

$$
\text{s.t.} \quad \left.\begin{aligned} &\sum_{h_\alpha\backslash h_i}d_{(x,y),\alpha}(h_\alpha) = d_{(x,y),i}(h_i) \quad \forall(x,y), i\in\mathbb{H}, \alpha\in N(i), h_i\in\mathcal{S}_i\\ &d_{(x,y),i}, d_{(x,y),\alpha}\in\underline{\Delta}\end{aligned}\right\} := d_{(x,y)}\in\mathcal{C}_{(x,y)} \quad \forall(x,y)\in\mathcal{D}
$$

---

**Claim 1** *The function*

$$
\frac{C}{p}\|w\|_p^p + \sum_{(x,y)}\left(\epsilon\ln\sum_{\hat{s}\in\mathcal{S}}\exp\left(\frac{w^\top\phi(x,\hat{s}) + \ell_{(x,y)}(\hat{s})}{\epsilon}\right) - \right.
$$
$$
\left. -\epsilon H(q_{(x,y)}) - \mathbb{E}_{q_{(x,y)}}[w^\top\phi(x,(y,\hat{h})) + \ell^c(x,(y,\hat{h}))]\right) \quad (6)
$$

*convex in $w$ and $q_{(x,y)}$ separately, is an upper bound on Eq. (5), $\forall q_{(x,y)}\in\underline{\Delta}$, with $\underline{\Delta}$ the probability simplex, $H$ the entropy and $\mathbb{E}$ the expectation w.r.t. the stated distribution. The bound holds with equality for that $q^*_{(x,y)}(h)$ minimizing this cost function (Eq. (6)).*

**Proof:** In supplementary material □

For many real-world applications, the program in Claim 1 involves sums over exponentially sized sets $\mathcal{S}$ and $\mathcal{H}$. They are exponentially sized as the observed and unobserved labels $y = (s_i)_{i\in\mathbb{Y}} \in \mathcal{Y}$ and $h = (s_i)_{i\in\mathbb{H}} \in \mathcal{H}$ are often tuples with elements $s_i\in\mathcal{S}_i$ taking $|\mathcal{S}_i|$ discrete states. Note that $\mathbb{S} = \mathbb{Y}\cup\mathbb{H}$, with product spaces $\mathcal{Y} = \prod_{i\in\mathbb{Y}}\mathcal{S}_i$ and $\mathcal{H} = \prod_{i\in\mathbb{H}}\mathcal{S}_i$. But the features usually describe interactions only between smaller subsets of random variables

$$
\phi_r(x,s) = \sum_{\alpha\in E_r}\phi_{r,\alpha}(x,s_\alpha) + \sum_{i\in\mathbb{S}_r}\phi_{r,i}(x,s_i), \quad (7)
$$

where $E_r$ and $\mathbb{S}_r$ denote the sets of factors and variables, and $\mathbb{S} = \bigcup_r \mathbb{S}_r$. Note that each feature is described by a bipartite factor graph $G_r$ with nodes originating from the variable set $\mathbb{S}_r$ and factors from $E_r$. An edge connects a single node $i\in\mathbb{S}_r$ to a factor $\alpha\in E_r$ iff $i\in\alpha$. Consider the factor graph $G = \bigcup_r G_r$ where we define the set of neighbors $N(i) := \{\alpha : i\in$ $\alpha, \; \forall\; \alpha\in E\}$ and $N(\alpha) := \{i : i\in\alpha, \; \forall\; i\in\mathbb{S}\}$. In many applications the loss functions $\ell$ and $\ell^c$ factorize in a similar fashion and are easily introduced in the graphical model $G$, *i.e.*, $\ell_{(x,y)}(s)$ decomposes into local terms $\ell_{(x,y),i}(s_i)$, $\forall i\in\mathbb{S}$ and interaction terms $\ell_{(x,y),\alpha}(s_\alpha)$ $\forall\alpha\in E$, whereas $\ell^c_{(x,y)}(\hat{h})$ is structured according to the locally defined variables $\ell^c_{(x,y),i}(\hat{s}_i)$, $\forall i\in\mathbb{H}$ and $\ell^c_{(x,y),\alpha}((y,\hat{h})_\alpha)$.

We make use of the local structure of features and loss, and approximate the intractable function in Claim 1. In particular, let the probability distribution $q_{(x,y)}(h)$ be described by local beliefs $d_{(x,y),i}(h_i)\in\underline{\Delta}$ and factor beliefs $d_{(x,y),\alpha}(h_\alpha)\in\underline{\Delta}$. We approximate the marginal polytope by a local one using the marginalization constraints $\sum_{h_\alpha\backslash h_i}d_{(x,y),\alpha}(h_\alpha) = d_{(x,y),i}(h_i)$ $\forall(x,y)\in\mathcal{D}, i\in\mathbb{H}, \alpha\in N(i), h_i\in\mathcal{S}_i$. We introduce counting numbers $\hat{c}_i$ and $\hat{c}_\alpha$ to allow for more flexibility in the approximation. To further obtain a tractable approximation for the partition function over $\mathcal{S}$ we approximate its Legendre transform, an entropy ranging over $s\in\mathcal{S}$, via local terms. As those local terms are required to fulfill marginalization constraints for global consistency in the dual domain, we obtain Lagrange multipliers $\lambda_{(x,y),i\to\alpha}(s_i)$ $\forall(x,y)\in\mathcal{D}$, $i\in\mathbb{S}$, $\alpha\in N(i)$ and $s_i\in\mathcal{S}_i$ on the graph $G$ for the primal formulation. Note that those Lagrange multipliers are often interpreted as messages. For generality of the entropy approximations we again allow for counting numbers $c_i$ and $c_\alpha$. We now formally state our approximation.

**Theorem 1** *The approximation of the program in Eq. (6) takes the form given in Program 1 where*

$\phi_{(x,y),i}(s_i) = \ell_{(x,y),i}(x, s_i) + \sum_{r:i \in \mathbb{S}_r} w_r \phi_{r,i}(x, s_i)$ *and*
$\phi_{(x,y),\alpha}(s_\alpha) = \ell_{(x,y),\alpha}(x, s_\alpha) + \sum_{r:\alpha \in E_r} w_r \phi_{r,\alpha}(x, s_\alpha)$.

**Proof:** In supplementary material □

## 4. Message Passing Algorithm

Before deriving an algorithm for solving Program 1, we begin by discussing the properties of the approximation. For counting numbers and annealing factor $\epsilon$ larger than zero, it is jointly convex in the messages $\lambda_{(x,y),i \to \alpha}(s_i)$ $\forall (x,y) \in \mathcal{D}, i \in \mathbb{S}, \alpha \in N(i), s_i \in \mathcal{S}_i$ and the model parameters $w$. It is also jointly convex in the messages and the beliefs $d_{(x,y),i}$ $\forall i \in \mathbb{H}$ and $d_{(x,y),\alpha}$ $\forall \alpha \in E_{\mathbb{H}}$, but not jointly convex when optimizing for both the weights and the beliefs. Cycling through blocks of variables and updating them in a block-coordinate descent manner is not guaranteed to converge as we cannot fulfill pseudoconvexity in every pair of coordinate blocks. Similar to other latent variable frameworks we can obtain convergence guarantees when employing instances of variational methods discussed, *e.g.*, in (Jordan et al., 1999) like the concave-convex procedure (CCCP) (Yuille & Rangarajan, 2003; Sriperumbudur & Lanckriet, 2009) by separating the cost function into two functions $f_1(w, \lambda)$ and $f_3(d)$, convex in their parameters and a bilinear term $f_2(w, d)$ connecting the two. We refer the reader to Program 1 for the definition of these functions. Here $\lambda$ is the vector of all messages, $d$ the vector of all beliefs, and $\mathcal{C}_{(x,y)}$ $\forall (x,y) \in \mathcal{D}$ the set of all marginalization constraints.

Without loss of generality we can assume Program 1 to be bounded from below. Considering the biconvex cost function, it is intuitive to alternate between solving for the beliefs and then performing a gradient step in the direction of the weights and the messages. Due to the fact that the program is unconstrained in the messages and model parameters, one gradient step of the latter is sufficient. We refer the reader to the supplementary material for a detailed derivation of the algorithm. In short, updating the beliefs, *i.e.*, the 'latent variable prediction problem' requires solving

$$\min_{d_{(x,y)}} f_2(w, d) + f_3(d) \quad \text{s.t.} \quad d_{(x,y)} \in \mathcal{C}_{(x,y)} \quad (8)$$

for every $(x,y) \in \mathcal{D}$ independently, hence possibly in parallel. This problem reduces to a standard (convex) belief propagation task (Hazan & Shashua, 2010) which is guaranteed to find the global optimum for strictly positive counting numbers $\hat{c}_i$, $\hat{c}_\alpha$ and annealing factor $\epsilon$. To update the weights and messages we are required to decrease the cost function of the fol-

---

**Algorithm 1** latent structured prediction

**repeat**
 **repeat**
  //to solve latent variable prediction problem
  $\min_d f_2 + f_3$ s.t. $\forall (x,y)$ $d_{(x,y)} \in \mathcal{D}_{(x,y)}$
 **until** convergence
 //message passing update
 $\forall (x,y), i \in \mathbb{S}$  $\lambda_{(x,y),i} \leftarrow \nabla_{\lambda_{(x,y),i}}(f_1 + f_2) = 0$
 //gradient step with step size $\eta$
 $w \leftarrow w - \eta \nabla_w (f_1 + f_2)$
**until** convergence

---

lowing unconstrained program, convex in $w$ and $\lambda$:

$$\min_{w, \lambda} f_1(w, \lambda) + f_2(w, d). \quad (9)$$

Similar to the program given in Eq. (8), convergence is guaranteed for counting numbers and annealing factor being strictly positive. More importantly, for weights $w$, one gradient step of length $\eta$ obtained via line search is sufficient for convergence guarantees. A solution for a block-coordinate descent step $\nabla_{\lambda_{(x,y),i}}(f_1 + f_2) = 0$ w.r.t. $\lambda_{(x,y),i \to \alpha}(s_i)$ for $(x,y) \in \mathcal{D}, i \in \mathbb{S}$ can be analytically computed jointly $\forall \alpha \in N(i), s_i$. We briefly state the proposed algorithm for latent structured prediction in Alg. 1 while pointing the interested reader to the supplementary material for details. Some convergence properties of the proposed algorithm are summarized in the following claim.

**Claim 2** *Alg. 1 is guaranteed to decrease the cost function of Program 1 at every iteration and guaranteed to converge to a minimum or a saddle point for $\epsilon, c_i, c_\alpha, \hat{c}_i, \hat{c}_\alpha > 0$.*

**Proof:** In supplementary material □

## 5. Experiments

In this section we demonstrate the effectiveness of our approach in the tasks of image segmentation as well as 3D scene understanding, and show that our method significantly outperforms LSSVM in terms of performance and HCRF in terms of speed.

**Segmentation:** Our first task addresses segmentation of weakly labeled images. This is an interesting example, as the graphical model contains many loops. As ground truth we use the $14 \times 40$ sized "ICML" tag given in Fig. 1. We created a dataset composed of 10 training and 10 test instances, where each observation $x$ is obtained by adding zero mean, uniform noise on the ground truth labels $y_i \in \mathcal{S}_i = \{1, \ldots, 5\}$. We employ $F = 2$ features, a local potential based on the observations and a pairwise linear smothness potential.

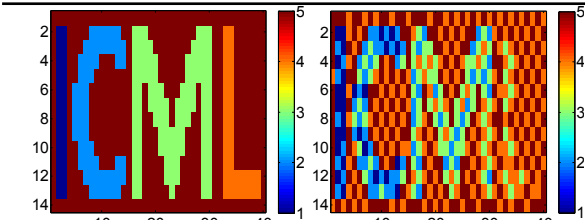

*Figure 1.* Segmenting the ICML tag with 90% latent variables: (left) ground truth and our result, (right) wrong model learned by LSSVM.

$G$ is a grid-like graph, typical for many vision applications. In our experiments, we gradually increase the amount of missing labels from 0% to 100%, and determine at random which variables are hidden/latent.

We compare our approach to a standard HCRF and the latent structured SVM (LSSVM) of Yu & Joachims (2009) which uses belief propagation to solve the respective sub-problems. We use at most 200 outer iterations, 1000 inner iterations for our approach, and 200 outer iterations, 1000 message passing iterations, 20 cutting plane iterations for LSSVM. For computational reasons the HCRF method is restricted to only 10 outer iterations, 1000 message passing iterations and 5 CRF iterations resulting in a maximum of 50 updates of the model parameters. For a fair comparison, we also use 50 outer iterations and 1000 message passing iterations resulting in a maximum of 50 updates for our approach with $\epsilon = 1$. All algorithms employed the same initialization. For our framework we additionally vary $\epsilon$ from 0 for the max-margin formulation to 1 for the maximum-likelihood formulation. Mean results averaged over 5 runs are depicted in Fig. 2. Our method results in good prediction for all values of $\epsilon$, while the LSSVM fails in the presence of large amounts of latent variables. This is due to the fact that the cutting planes are not exactly computable for loopy models ($(14 \cdot 40)^5$ possibilities), and thus no decrease in the cost function is guaranteed. An example of the prediction of our approach and LSSVM in the presence of 90% latent variables is illustrated in Fig. 1, where LSSVM learns a wrong model that favors neighboring pixels to be different. In contrast, the HCRF performs similarly to our approach, but it takes on average 213.2min to compute a single HCRF experiment while only 6.2min are required for our approach with $\epsilon = 1$. Since HCRF is not practical, we focus the rest of the experimental evaluation on LSSVM.

**3D Scene Understanding:** Recovering the spatial layout of indoor scenes from a single image is an important problem in applications such as personal robotics and computer vision. Existing approaches formulate the problem as a structured prediction task focusing on estimating the 3D box which best describes the scene layout. Taking advantage of the *Manhat-*

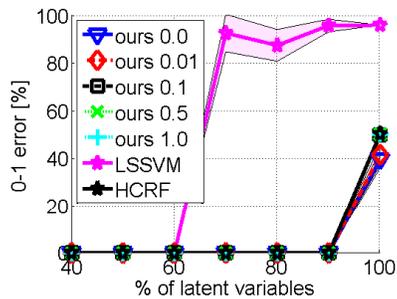

*Figure 2.* Performance as a function of the amount of latent variables, averaged over 5 runs.

*tan world assumption* (*i.e.*, there exist three dominant vanishing points which are orthonormal), the problem can be formulated as inference in a fully connected pairwise graphical model $G$ composed of four random variables. As shown in Fig. 3(a), these variables represent the angles encoding the rays that originate from the respective vanishing points. Following existing approaches (Hedau et al., 2009; Lee et al., 2010), we employ $F = 55$ features based on geometric context (GC) and orientation maps (OM) and refer the interested reader to (Hoiem et al., 2007) and (Lee et al., 2009) for respective details. Our features count for each face in the cuboid (given a particular configuration of the layout) the number of pixels with a certain label for OM and the probability that such label exists for GC. Performance is measured as the percentage of pixels that have been correctly labeled, with labels, *i.e.*, left-wall, right-wall, front-wall, ceiling, floor.

We first investigate how the layout estimation can benefit from the use of weakly labeled data. To this end we use a set of fully annotated images, denoted 'fixed,' and add a varying number {25, 50, 100} of images with only 1 or 2 angles labeled, *i.e.*, 75% or 50% missing information. The randomly chosen unlabeled angles are treated as latent variables. All results are averaged over 12 runs, each being trained on a varying portion of the training set. Learning is performed with parameters $C = 1$, $\epsilon = 0.01$ and all counting numbers equal to one. The results for 50% and 75% of missing information are detailed in Fig. 3(b) and Fig. 3(c) respectively. As expected, the prediction performance improves as a function of the number of fully labeled images, but more importantly, the performance also significantly improves as a function of the amount of weakly labeled data. Our performance also increases as a function of how much supervision the weakly annotated images have, *i.e.*, 2 hidden variables outperforms having 3 latent variables per image.

In the next experiment we compare our approach to LSSVM. Again, all results are averaged over 12 runs. Note that we have to modify the stopping criteria of LSSVM as we are not guaranteed to find decreasing steps at each iteration. In the absence of any clear criterion, we force LSSVM to perform at least 10 outer loops. Fig. 4 shows results when adding 25 (column 1), 50 (column 2) or 100 (column 3) weakly labeled examples with 50% (row 1) or 75% (row 2) missing information. Our approach significantly outperforms LSSVM in all settings.

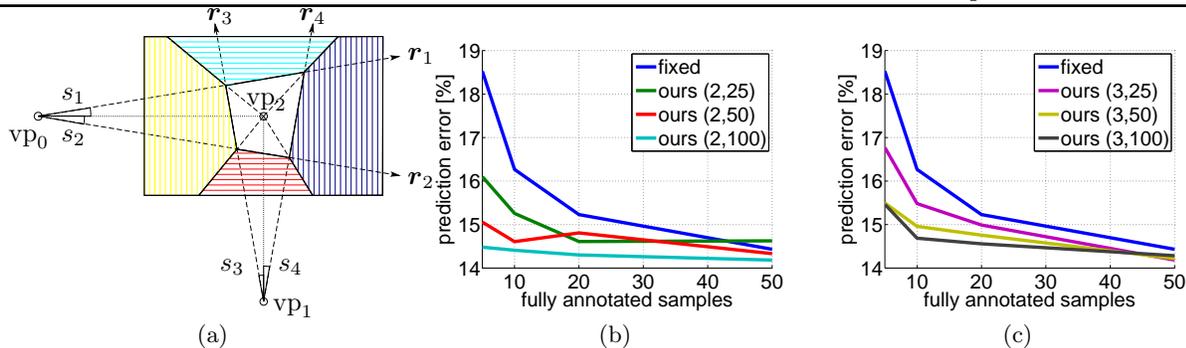

*Figure 3.* The parameterization of the 3D scene understanding task is illustrated in (a). Comparing different amount ({25, 50, 100}) of weakly labeled additional information with 50% and 75% missing data in (b) and (c).

| | fully | weakly | Error |
|---|---|---|---|
| (Hoiem et al., 2007) | 209 | 0 | 28.9% |
| (Hedau et al., 2009) | 209 | 0 | 21.2% |
| (Wang et al., 2010) | 209 | 0 | 22.2% |
| (Lee et al., 2010) | 209 | 0 | 18.6% |
| ours | 10 | 25 (75%) | **15.5%** |
| ours | 10 | 50 (75%) | **15.0%** |
| ours | 10 | 100 (75%) | **14.7%** |

*Table 1.* Comparison to state-of-the-art on the layout data set of (Hedau et al., 2009). 75% of the information is missing for each weakly annotated image.

A comparison of our approach to the state-of-the-art is shown in Tab. 1. Great performance is achieved with a small amount of supervision. Our fully supervised approach with 200 completely labeled examples results in a prediction error of 13.6% (Schwing et al., 2012).

Fig. 5 depicts improvements achieved by our approach compared to only using the fixed set of 10 fully labeled training images, as well as LSSVM. For LSSVM and our approach, we used an additional 100 images with 50% missing annotations. Prediction errors are indicated below the figures. We also provide illustrations for the image features we employed in the last two columns, *i.e.*, orientation maps and geometric context. Interestingly, when trained with only 10 images, the model tends to miss walls and the ceiling.

## 6. Related Work and Discussion

HCRFs (Quattoni et al., 2007) and LSSVMs (Yu & Joachims, 2009) are the most common frameworks employed to deal with latent variable models in structured prediction problems. The first contribution of our work described in Sec. 2 and formalized in Eq. (5) is to unify the aforementioned frameworks. More specifically, our max-margin formulation ($\epsilon = 0$) is identical to the formulation presented by Yu & Joachims (2009) when having $p = 2$, $\ell^c \equiv 0 \; \forall x, h, \hat{h}$. The weak-label structured SVM presented in (Girshick et al., 2011) is obtained when $\epsilon = 0$ and $p = 2$. For $\epsilon = 1$, $p = 2$, $\ell^c \equiv 0 \; \forall x, y, \hat{h}$ and $\ell \equiv 0 \; \forall x, y, \hat{y}, \hat{h}$ we recover the likelihood formulation presented by Quattoni et al. (2007). For general $\epsilon$, but without latent variables, *i.e.*, $\mathcal{H} = \varnothing \; \forall (x, y)$, our formulation reduces to the one presented in (Hazan & Urtasun, 2010) which generalizes structured SVMs and CRFs. Importantly, through the $\epsilon$ parameter our work introduces a family of new latent variable models in structured prediction that range between HCRF and LSSVM.

The main drawback of previous works (Quattoni et al., 2007; Yu & Joachims, 2009) is that they rely on computing the MAP estimate or the partition function at each iteration. In the case of general graphical models, approximate inference techniques like belief propagation are employed. The influence of approximate inference algorithms on structured SVMs (Taskar et al., 2004; 2005; Tsochantaridis et al., 2004) without latent variables has been investigated by (Finley & Joachims, 2008; Kulesza & Pereira, 2008), where they reported a "generally poor performance" when combining belief propagation and structured SVMs. As an LSSVM approach employs a structured SVM in every iteration, we expect a similar behavior when combining LSSVM with belief propagation. This was indeed the conclusion of our experiments in Sec 5. Similar to (Finley & Joachims, 2008), we found that ties within the solution mislead LSSVM.

To address efficiency Komodakis (2011) suggested to use a small number of CRF iterations. This, however, would not have convergence guarantees. Our second contribution, detailed in Sec. 3, is to directly include the approximation into the cost function. As a result we are able to derive a message passing algorithm that

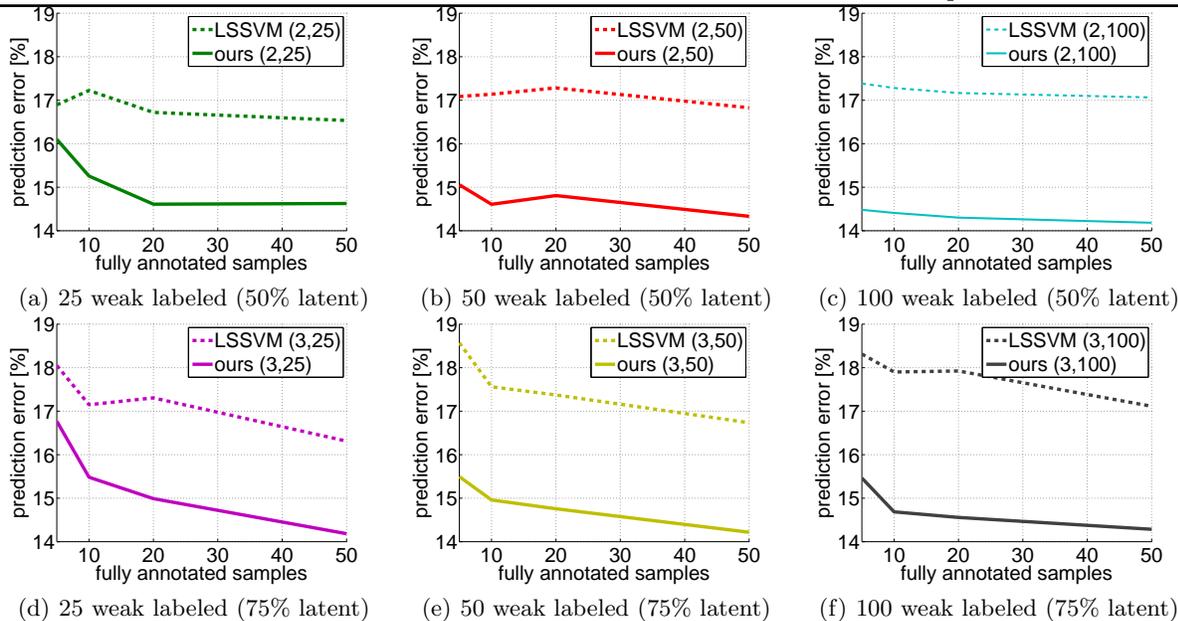

(a) 25 weak labeled (50% latent)  (b) 50 weak labeled (50% latent)  (c) 100 weak labeled (50% latent)

(d) 25 weak labeled (75% latent)  (e) 50 weak labeled (75% latent)  (f) 100 weak labeled (75% latent)

*Figure 4.* Comparison of our approach with LSSVM, where we measure pixel-wise prediction performance when adding 25 (column 1), 50 (column 2) or 100 (column 3) weakly labeled data with 50% (row 1) or 75% (row 2) missing information.

is significantly more efficient and guaranteed to converge. Our method needs to solve the 'latent variable prediction problem' just like LSSVM or HCRF. However, LSSVM and HCRF also require to solve the loss-augmented inference problem in every iteration before performing a parameter update. In contrast, we only need a single update on the messages $\lambda$ before updating $w$. This results in large speedups as demonstrated in the previous section.

In our experiments we also observe that the loss function is very important when learning from weakly labeled data. In HCRFs, no loss function was proposed. For LSSVMs, the standard structured SVM loss was applied and adapted by Komodakis (2011). Girshick et al. (2011) proposed to introduce a second loss function into the 'latent variable prediction problem' while Tarlow & Zemel (2012) investigate the impact of higher order loss functions. Kumar et al. (2010) proposed the self paced learning algorithm, which starts with "easy" examples before gradually adding more difficult ones. Their formulation is based on LSSVMs but can also be applied to our framework.

Our algorithm is easily parallelized w.r.t. to the data samples. Our C++ implementation uses OpenMP and MPI for parallelization in both shared and distributed memory environments. To parallelize message passing one could employ (Schwing et al., 2011). The sources are available on `http://alexander-schwing.de`.

## 7. Conclusion

We have proposed a framework that unifies HCRF and latent structured SVMs. We have then constructed an approximation of the resulting intractable optimization problem using local entropies, and derived an algorithm for general graphs that leverages the graphical model structure imposed by the features. We have demonstrated the effectiveness of our approach on a segmentation task as well as predicting the 3D layout from single images. We plan to extend this work in two directions along the lines of v.d.Maaten et al. (2011), by addressing non-linear structured prediction with latent variables and by investigating relations to deep belief networks.

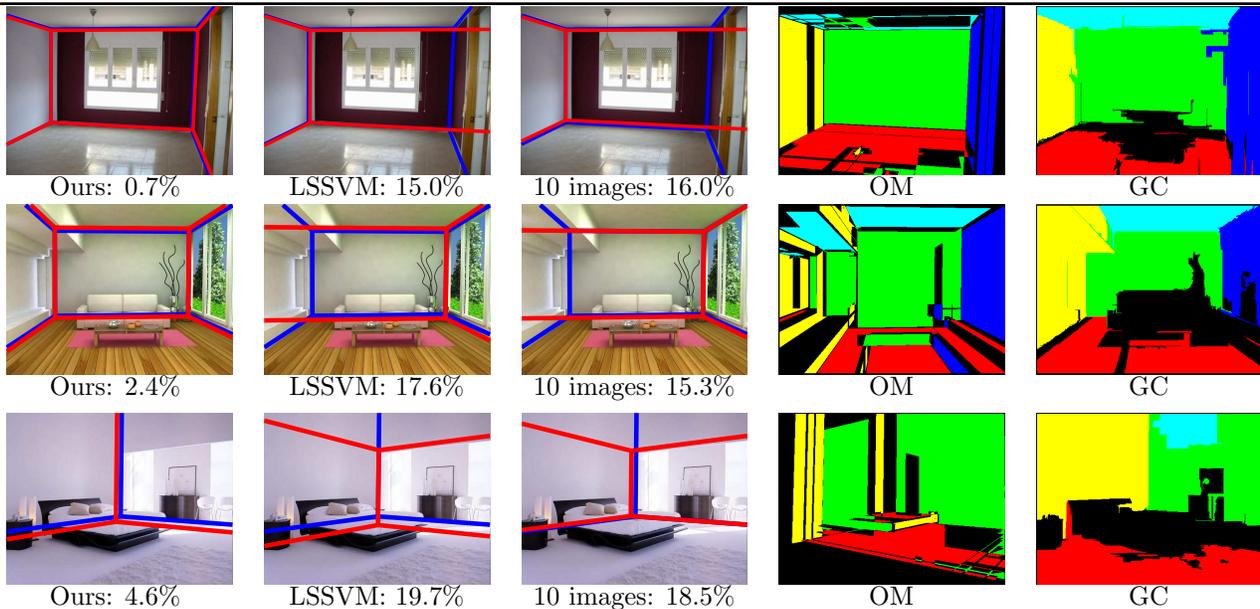

*Figure 5.* Column 1 and 2 show prediction results of our approach and LSSVM after training with 10 fully labeled images and an additional 100 images labeled with 50% missing information. Column 3 illustrates prediction results after training with only 10 fully labeled images. The pixel errors are as indicated below each figure. Column 4 and 5 provide the orientation map and geometric context image features.